\documentclass[10pt,twocolumn,letterpaper]{article}

\usepackage{iccv}
\usepackage{times}
\usepackage{epsfig}
\usepackage{graphicx}
\usepackage{amsmath}
\usepackage{amssymb}
\usepackage{multirow}
\usepackage{threeparttable}
\usepackage[table]{xcolor}
\usepackage[super]{nth}
\usepackage{arydshln}
\usepackage{xcolor}
\usepackage[export]{adjustbox}
\usepackage[pagebackref=true,breaklinks=true,letterpaper=true,colorlinks,bookmarks=false]{hyperref}
\iccvfinalcopy %

\def\iccvPaperID{2099} %

\begin{document}

\title{Corners for Layout: End-to-End Layout Recovery from 360 Images}

\author{Clara Fernandez-Labrador\textsuperscript{1,2,*}\quad Jose M. Facil\textsuperscript{1,*}\quad Alejandro Perez-Yus\textsuperscript{1}\quad \\
{\tt\small cfernandez@unizar.es} \quad{\tt\small jmfacil@unizar.es}\quad  {\tt\small alperez@unizar.es}\\
C{\'e}dric Demonceaux\textsuperscript{2}\quad Javier Civera\textsuperscript{1}\quad Jose J. Guerrero\textsuperscript{1} \\
{\tt\small cedric.demonceaux@u-bourgogne.fr} \quad {\tt\small jcivera@unizar.es} \quad {\tt\small josechu.guerrero@unizar.es}\\
\textsuperscript{1} University of Zaragoza\quad \textsuperscript{2} Universit\'e Bourgogne Franche-Comt\'e \\
}

\maketitle
\renewcommand*{\thefootnote}{\fnsymbol{footnote}}
\footnotetext[1]{Equal contribution}
\newcommand{\shortMethod}{CFL}
\newcommand{\longMethod}{Corners for Layout}
\newcommand{\stb}{\textit{smaller is better}}
\newcommand{\btb}{\textit{bigger is better}}

\newcommand{\sbt}{\,\begin{picture}(-1,1)(-1,-3)\circle*{3}\end{picture}\ }

\newcommand{\shortConvs}{EquiConvs}
\newcommand{\longConvs}{Equirectangular Convolutions}

\newcommand{\shortSphericalK}{$K_{\mathcal{S}ph}$}
\newcommand{\longSphericalK}{spherical surface kernel}

\definecolor{cAlphas}{HTML}{4444CC}%
\definecolor{cRes}{HTML}{E00000} %
\definecolor{cTheta}{HTML}{2CA02C} %
\definecolor{cPhi}{HTML}{FF871C} %

\newcommand{\todo}[1]{{\scriptsize \color{blue}{#1}}}

\definecolor{magenta}{HTML}{FF00FF} %
\definecolor{dmagenta}{HTML}{7A007A}

\newcommand\Tstrut{\rule{0pt}{2.6ex}}         %
\newcommand\Bstrut{\rule[-0.9ex]{0pt}{0pt}}   %

\definecolor{trasla}{HTML}{1E90FF} %
\definecolor{rota}{HTML}{FF8C00} %
\begin{abstract}

The problem of 3D layout recovery in indoor scenes has been a core research topic for over a decade. However, there are still several major challenges that remain unsolved. Among the most relevant ones, a major part of the state-of-the-art methods make implicit or explicit assumptions on the scenes --\eg box-shaped or Manhattan layouts. Also, current methods are computationally expensive and not suitable for real-time applications like robot navigation and AR/VR.
In this work we present CFL (Corners for Layout), the first end-to-end model for 3D layout recovery on $360^\circ$ images. Our experimental results show that we outperform the state of the art, making less assumptions on the scene than other works, and with lower cost. We also show that our model generalizes better to camera position variations than conventional approaches by using \shortConvs{}, a convolution applied directly on the spherical projection and hence invariant to the equirectangular distortions.

\vspace{-4mm}
\end{abstract}
\section{Introduction}

Recovering the 3D layout of an indoor scene from a single view has attracted the attention of computer vision and graphics researchers in the last decade.
The idea is going beyond pure geometrical reconstructions and provide higher-level contextual information about the scene, even in the presence of clutter.
Layout estimation is a key technology in several emerging application markets, such as augmented and virtual reality and robot navigation. But also for more traditional ones, like real estate \cite{Liu_2015_CVPR}.%

Layout estimation, however, is not a trivial task and there are several major problems that still remain unsolved. For example, most existing methods are based on strong assumptions on the geometry (\eg Manhattan scenes) or the over-simplification of the room types (\eg box-shaped layouts), often underfitting the richness of real indoor spaces. The limited field of view of conventional cameras leads to ambiguities, which could be solved by considering a wider context. For this reason it is advantageous to use wide fields of view, like 360$^\circ$ panoramas. In these cases, however, the methods for conventional cameras are not suitable due to the image distortions and new ones have to be developed.

In the last years, the main improvements in layout recovery from panoramas have come from the application of deep learning. The high-level features learned by deep networks have proven to be as useful for this problem as for many others. Nevertheless, these techniques entail other problems such as the lack of data or overfitting. State-of-the-art methods require additional pre- and/or post-processing. As a consequence they are very slow, and this is a major drawback considering the aforementioned applications for real-time layout recovery.

\begin{figure}
\centering
\includegraphics[width=1.\linewidth]{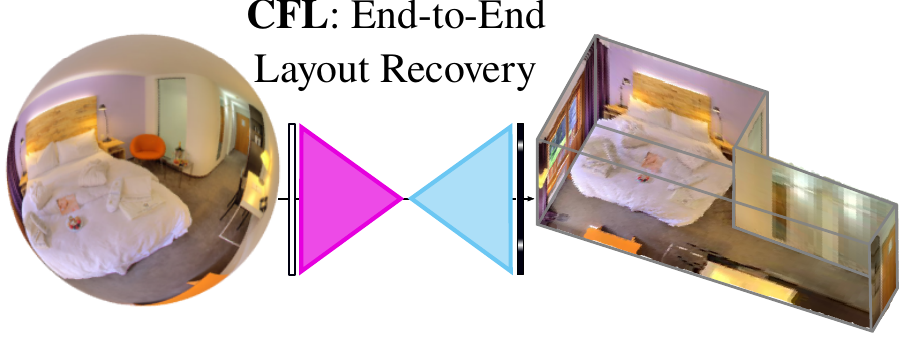}
\caption{\label{fig:teaserIM}\textbf{\longMethod{}}: The first end-to-end model from the sphere to the 3D layout.}
\vspace{-4mm}
\end{figure}

In this work, we present \longMethod{} (\shortMethod{}) the first end-to-end neural network that recovers the 3D layout from a single $360^\circ$ image (Figure~\ref{fig:teaserIM}).
\shortMethod{} predicts a map of the corners of the room that is directly used to obtain the layout without further processing. %
This makes \textbf{\shortMethod{} more than 100 times faster} than the state of the art, while still \textbf{outperforming the accuracy of current approaches}.
Furthermore, our proposal is not limited by typical scene assumptions, meaning that it can predict complex geometries, such as rooms with more than four walls or non-Manhattan structures. 
Additionally, we propose a novel implementation of the convolution for $360^\circ$ images \cite{tateno2018distortion,cohen2018spherical} in the 
equirectangular projection. %
We deform \cite{dai2017deformable} the kernel to compensate the distortion and make \shortMethod{} more \textbf{robust to camera rotation and pose variations}. 
Hence, it is equivalent to applying directly a convolution operation to the spherical image, which is geometrically more coherent than applying a standard convolution on the equirectangular panorama.
We have extensively evaluated our network in two public datasets with several training configurations, including data augmentation techniques to address occlusions by enforcing the network to learn from the context. %
Our \textbf{code} and labeled \textbf{dataset} can be found here: \href{https://cfernandezlab.github.io/CFL/}{CFL webpage}.

\section{Related Work}

The layout of a room provides a strong prior for other visual tasks like depth recovery \cite{Eigen2015}, realistic insertions of virtual objects into indoor images \cite{karsch2011rendering}, indoor object recognition \cite{bao2011toward,song2016deep} or human pose estimation \cite{fouhey2014people}. A large variety of methods have been developed for this purpose using multiple input images \cite{tsai2011real,flint2011manhattan} or depth sensors \cite{zhang2013estimating}, which deliver high-quality reconstruction results. For the common case when a single RGB image is available, the problem becomes considerably more challenging and researchers need very often to rely on strong assumptions. %

The seminal approaches to layout prediction from a single view were \cite{Delage2006,Lee2009}, followed by \cite{Hedau2009,schwing2013box}. They basically model the layout of the room with a vanishing-point-aligned 3D box, being hence constrained to this particular room geometry and unable to generalize to others appearing frequently in real applications.
Most recent approaches exploit CNNs and their excellent performance in a wide range of applications such as image classification, segmentation and detection. \cite{Mallya:2015,ren2016coarse,Mallya2,zhao2017physics}, for example, focus on predicting the informative edges separating the geometric classes (walls, floor and ceiling). Alternatively, Dasgupta \etal \cite{dasgupta2016delay} proposed a FCN to predict labels for each of the surfaces of the room. All these methods require extra computation added to the forward propagation of the network to retrieve the actual layout. In \cite{RoomNet}, for example, an end-to-end network predicts the layout corners in a perspective image, but after that it has to infer the room type within a limited set of manually chosen configurations. %

While layout recovery from conventional images has progressed rapidly with both geometry and deep learning, the works that address these challenges using omnidirectional images are still very few. 
Panoramic cameras have the potential to improve the performance of the task: their 360$^\circ$ field of view captures the entire viewing sphere surrounding its optical center, allowing to acquire the whole room at once and hence predicting layouts with more visual information. PanoContext \cite{PanoContext} was the first work that extended the frameworks designed for perspective images to panoramas. It recovers both the layout, which is also assumed as a simple 3D box, and bounding boxes for the most salient objects inside the room. Pano2CAD \cite{Pano2cad} extends the method to non-cuboid rooms, but it is limited by its dependence on the output of object detectors. Motivated by the need of addressing complex room geometries, \cite{fernandez2018layouts} generates layout hypotheses by geometric reasoning from a small set of structural corners obtained from the combination of geometry and deep learning. The most recent works along this line are LayoutNet \cite{zou2018layoutnet}, that trains a FCN from panoramas and vanishing lines, generating the layout models from edge and corner maps, and DuLa-Net \cite{yang2018dula}, that predicts Manhattan-world layouts leveraging a perspective ceiling-view of the room. All of these approaches require pre- or post-processing steps like line and vanishing point extraction or room model fitting, that increase their cost. 

In addition to all the challenges mentioned above, we also notice that there is an incrongruence between panoramic images and conventional CNNs. The space-varying distortions caused by the equirectangular representation makes the translational weight sharing ineffective. 
Very recently, Cohen \etal \cite{cohen2018spherical} did a relevant theoretical contribution by studying convolutions on the sphere using spectral analysis. However, it is not clearly demonstrated whether Spherical CNNs can reach the same accuracy and efficiency on equirectangular images. Our EquiConvs have more in common with the idea of \cite{tateno2018distortion}, that proposes distortion-aware convolutional filters to train their model using conventional perspective images and then use it to regress depth from panoramic images. We propose a novel parameterization and implementation of the deformable convolutions \cite{dai2017deformable} by following the idea of adapting the receptive field of the convolutional kernels by deforming their shape according to the distortion of the equirectangular projection.

\section{Corners for Layout}
Here we describe our end-to-end approach for recovering the layout, \textit{i.e.} the main structure of the room, from single $360^\circ$ images. After introducing some details about the target data, we describe the proposed network architecture and how we directly transform the output into the 3D layout. The network architecture is adapted for Standard Convolutions and for our proposed \longConvs{} implementation, the latest being explained in Section~\ref{sectionequiconv}.

\begin{figure}
\centering
\includegraphics[width=0.98\linewidth]{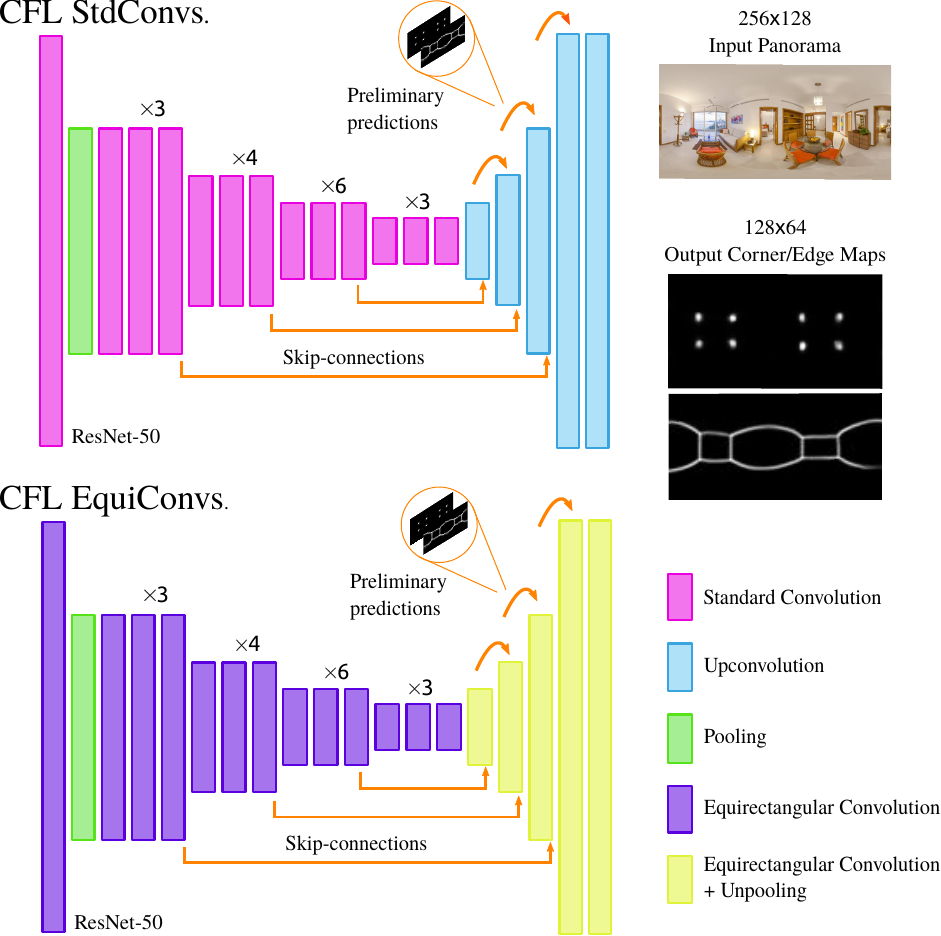}
\caption{\textbf{CFL architecture}. Our network is built upon ResNet-50, adding a single decoder that jointly predicts edge and corner maps. Here we propose two network variations: on top, the network applies StdConvs on the equirectangular panorama, whereas the one on the bottom applies \shortConvs{} directly on the sphere.}
\label{fig:Archit}
\vspace{-3mm}
\end{figure}

\subsection{Ground truth}
The ground truth (GT) for every panorama consists of two maps, $m$, one represents the room edges ($m=e$), \ie intersections between walls, ceiling and floor, and the other encodes the corner locations ($m=c$). Both maps are defined as $\mathcal{Y}^m=\{y_1^m, \hdots, y_i^m, \hdots\}$, with pixel values $y_i^m\in\left\{0,1\right\}$. $y_i^m$ has a value of $1$ if it belongs to an edge or a corner, and $0$ otherwise. We do line thickening and Gaussian blur for easier convergence during training since it makes the loss progression continuous instead of binary. The loss is gradually reduced as the prediction approaches the target.

Notice here that our target is considerably simpler than others that usually divide the ground truth into different classes. This contributes to the small computational footprint of our proposal. For example, \cite{Mallya:2015,zhao2017physics} use independent feature maps for background, wall-floor, wall-wall and wall-ceiling edges. A full image segmentation into left, front and right wall, ceiling and floor categories is performed in \cite{dasgupta2016delay}. In \cite{RoomNet}, they represent a total of 48 different corner types by a 2D Gaussian heatmap centered at the true keypoint location. Here, instead, we only use two probability maps, one for edges and another one for corners -- see \textit{outputs} in the Figure~\ref{fig:Archit}.

\subsection{Network architecture}
The proposed FCN follows the encoder-decoder structure and builds upon ResNet-50 \cite{he2016deep}. We replace the final fully-connected layer with a decoder that jointly predicts layout edges and corners locations already refined. We illustrate the proposed architecture in Figure~\ref{fig:Archit}.

\textbf{Encoder.}
Most of deep-learning approaches facing layout recovery problem have made use of the VGG16 \cite{simonyan2014very} as encoder \cite{Mallya:2015,dasgupta2016delay,RoomNet}. Instead, \cite{zhao2017physics} builds their model over ResNet-101 \cite{he2016deep} outperforming the state of the art. Here, we use ResNet-50 \cite{he2016deep}, pre-trained on the ImageNet dataset \cite{russakovsky2015imagenet}, which leads to a faster convergence due to the general low-level features learned from ImageNet. Residual networks allow us to increase the depth without increasing the number of parameters with respect to their plain counterparts. This leads, in ResNet-50, to capture a receptive field of $483 \times 483$ pixels, enough for our input resolution of $256 \times 128$ pixels. 

\textbf{Decoder.}
Most of the recent work \cite{Mallya:2015,zou2018layoutnet,ren2016coarse} builds two output branches for multi-task learning, which increases the computation time and the network parameters. We instead propose a unique branch with two output channels, corners and edge maps, which helps to reinforce the quality of both map types.
In the decoder, we combine two different ideas. First, skip-connections \cite{ronneberger2015u} from the encoder to the decoder. Specifically, we concatenate ``up-convolved'' features with their corresponding features from the contracting part. Second, we do preliminary predictions at lower resolutions which are also concatenated and fed back to the network following the spirit of \cite{dosovitskiy2015flownet}, ensuring early stages of internal features aim for the task.
We use ReLU as non-linear function except for the prediction layers, where we use Sigmoid.

We propose two variations of the network architecture for two different convolution operations (Figure~\ref{fig:Archit}).
 The first one, \textit{CFL StdConvs}, convolves the feature maps with Standard Convolutions and use up-convolutions to decode the output. The second one, \textit{CFL EquiConvs}, uses \longConvs{} both in the encoder and the decoder, using unpooling to upsample the output.
 \longConvs{} are deformable convolutions that adapt their size and shape depending on the position in the equirectangular image, for which we propose a new implementation in Section~\ref{sectionequiconv}.

\begin{figure}
\centering
\includegraphics[width=1\linewidth]{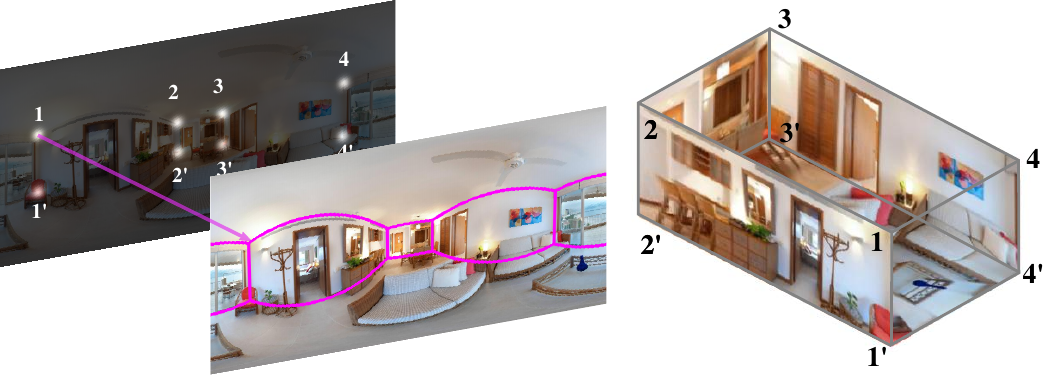}
\caption{\textbf{Layout from corner predictions}. From the corner probability map, the coordinates with maximum values are directly selected to generate the layout.}
\label{fig:Layout}
\vspace{-3mm}
\end{figure}

\subsection{Loss functions}
Edge and corner maps are learned through a pixel-wise sigmoid cross-entropy loss function. Since we know a priori that the natural distribution of pixels in these maps is extremely unbalanced ($\thicksim95\%$ have a value of $0$), we introduce weighting factors to make the training stable. Defining as $1$ and $0$ the positive and negative labels, the weighting factors are defined as $w_t = \frac{N}{N_t}$, being $N$ the total number of pixels and $N_t$ the amount of pixels of class $t$ per sample. The per-pixel per-map loss $\mathcal{L}_{i}^m$ is as follows:
\begin{eqnarray}
\mathcal{L}_{i}^m  &=& 
 w_1 \big( y_i^m\big(-\log(\hat{y}_i^m)\big)\big) + \nonumber \\
 & + & w_0 \big( (1-y_i^m) \big(-\log(1-\hat{y}_i^m)\big) \big),
\end{eqnarray}
\noindent where $y_i^m$ is the GT for pixel $i$ in the map $m$ and $\hat{y}_i^m$ is the network output for pixel $i$ and map $m$. We minimize this loss at $4$ different resolutions $k=\{1, \hdots, 4\}$, specifically in the network output ($k=4$) and 3 intermediate layers ($k=\{1, \hdots, 3\}$). The total loss is then the sum over all pixels, the $4$ resolutions and both the edge and corner maps
\begin{eqnarray}
\mathcal{L}  &=& \sum_{k=\{1, \hdots, 4\}} \sum_{m=\{e,c\}} \sum_{i} \mathcal{L}_{i}^m\left[k\right].
\end{eqnarray}

\subsection{3D Layout}

Aiming to a fast end-to-end model, CFL avoids post-processing and strong scene assumptions and just follow a natural transformation from corners coordinates to 2D and 3D layout. The 2D corners coordinates are the maximum activations in the probability map. Assuming that the corner set is consistent, they are directly joined, from left to right, in the unit sphere space and re-projected to the equirectangular image plane. From this 2D layout, we infer the 3D layout by only assuming ceiling-floor parallelism, leaving the wall structure unconstrained --\textit{i.e.}, we do not force the usual Manhattan perpendicularity between walls. Corners are projected to floor and ceiling planes given a unitary camera height (trivial as results are up to scale). See Figure~\ref{fig:Layout}.

\begin{figure}
\centering
    \includegraphics[width=0.77\linewidth]{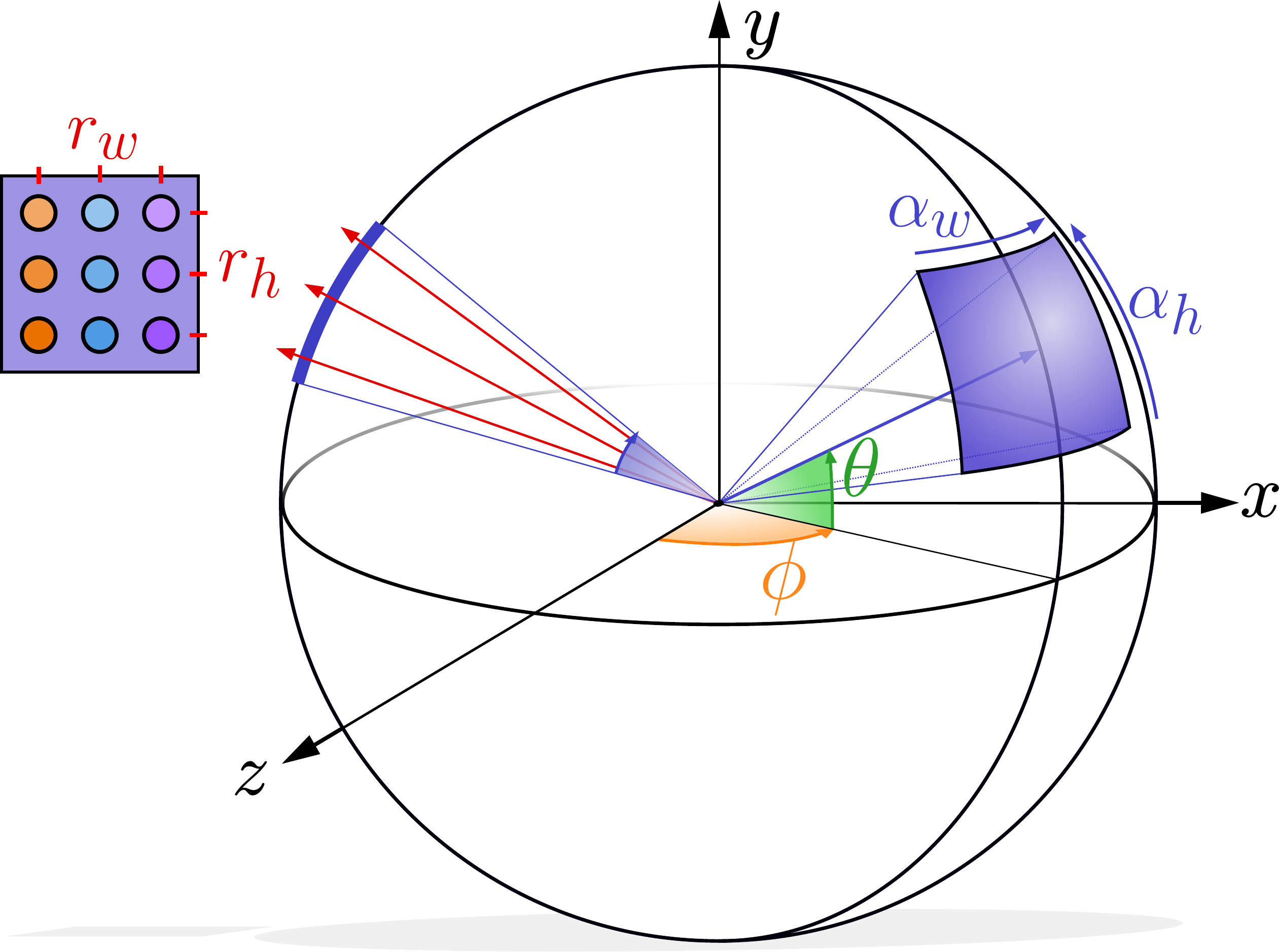}
\caption{\label{fig:equiconv} \textbf{Spherical parametrization of \shortConvs{}}. The spherical kernel, defined by its  angular size ({\color{cAlphas}$\alpha_w \times \alpha_h$}) and resolution ({\color{cRes}$r_w \times r_h$}), is convolved around the sphere with angles {\color{cPhi}$\phi$} and {\color{cTheta}$\theta$}.}
\vspace{-3mm}
\end{figure}

\noindent\textbf{Limitations of CFL:}
We directly join corners from left to right, meaning that our end-to-end model would not work if any wall is occluded because of the convexity of the scene. In those particular cases, the joining process should follow a different order. \cite{fernandez2018layouts} proposes a geometry-based post-processing that could alleviate this problem, but its cost is high and it needs the Manhattan World assumption. The addition of this post-processing into our work, in any case, could be done similarly to \cite{fernandez2018panoroom}. 

\section{\longConvs{}}
\label{sectionequiconv}

Spherical images are receiving an increasing attention due to the growing number of omnidirectional sensors in drones, robots and autonomous cars.
A na\"ive application of convolutional networks to a equirectangular projection, is not, in principle, a good choice due to the space-varying distortions introduced by such projection.

In this section we present a convolution that we name EquiConv, which is defined in the spherical domain instead of the image domain and it is implicitly invariant to equirectangular representation distortions. The kernel in \shortConvs{} is defined as a spherical surface patch --see Figure~\ref{fig:equiconv}. We parametrize its receptive field by the angles {\color{cAlphas}$\alpha_w$} and {\color{cAlphas}$\alpha_h$}. Thus, we directly define a convolution over the field of view. The kernel is rotated and applied along the sphere and its position is defined by the spherical coordinates ({\color{cPhi}$\phi$} and {\color{cTheta}$\theta$} in the figure) of its center. Unlike standard kernels, that are parameterized by their size $k_w \times k_h$, with \shortConvs{} we define the angular size ({\color{cAlphas}$\alpha_w \times \alpha_h$}) and resolution ({\color{cRes}$r_w \times r_h$}). In practice, we keep the aspect ratio, $\frac{{\color{cAlphas}\alpha_w}}{\color{cRes}r_w}=\frac{{\color{cAlphas}\alpha_h}}{\color{cRes}r_h}$, and we use square kernels, so we will refer the field of view as {\color{cAlphas}$\alpha$} ({\color{cAlphas}$\alpha_w=\alpha_h$}) and the resolution as {\color{cRes}$r$} ({\color{cRes}$r_w=r_h$}) respectively from now on. 
\begin{figure}
\centering
\includegraphics[width=1\linewidth]{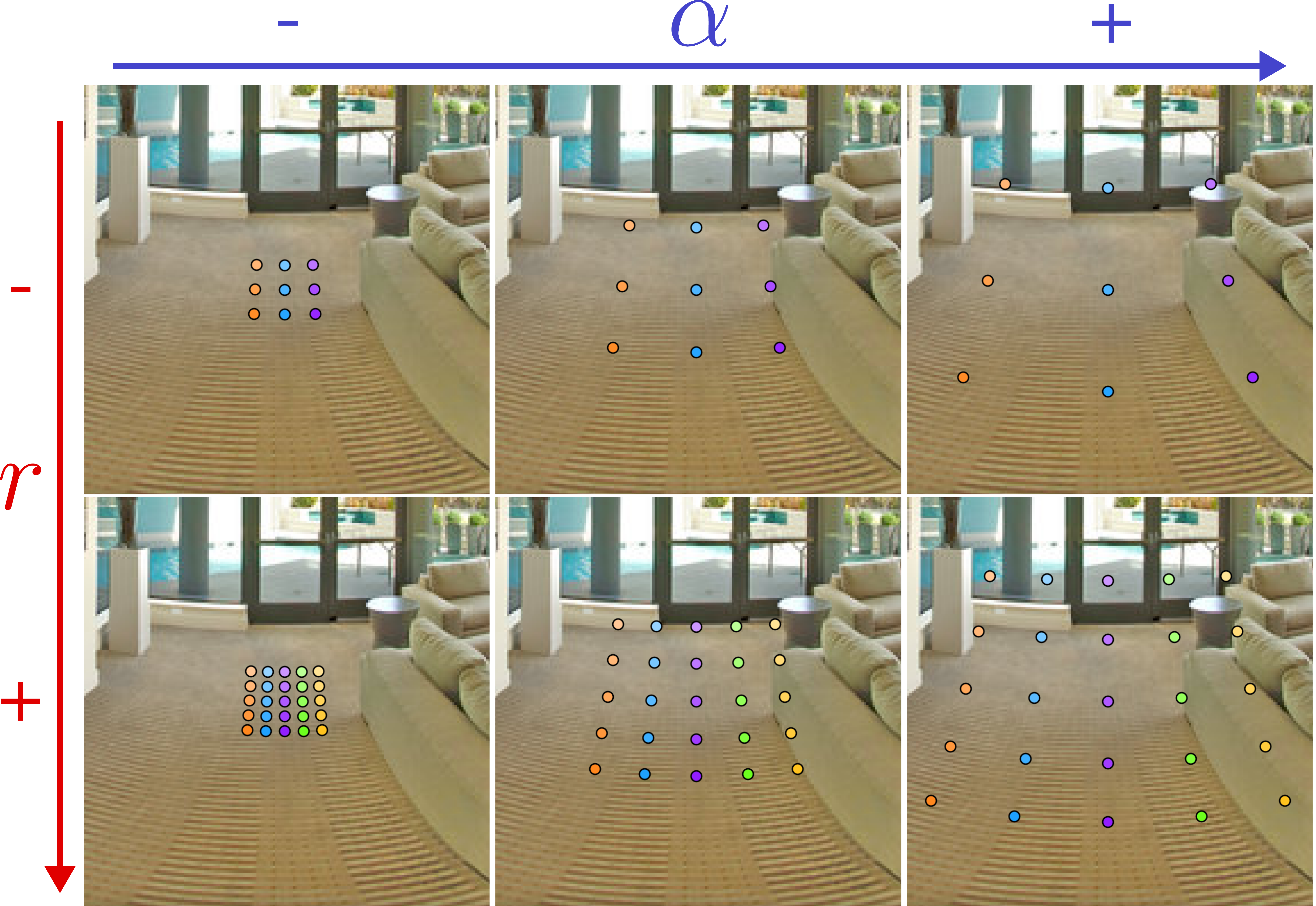}
\caption{\label{fig:sphericalatrous} \textbf{Effect of changing field of view {\color{cAlphas}$\alpha$} (rad) and resolution {\color{cRes}$r$} in \shortConvs{}}. \nth{1} column shows a narrow field of view {\color{cAlphas}$\alpha=0.2$}. \nth{2} column shows a wider kernel keeping its resolution (atrous-like), {\color{cAlphas}$\alpha=0.5$}. \nth{3} column shows an even larger field of view for the kernel, {\color{cAlphas}$\alpha=0.8$}. Notice how the kernel adapts to the equirectangular distortion. Rows are resolutions {\color{cRes}$r=3$} and {\color{cRes}$r=5$}.}
\vspace{-5mm}
\end{figure}
As we increase the resolution of the kernel, the angular distance between the elements decreases, with the intuitive upper limit of not giving more resolution to the kernel than the image itself. In other words, the kernel is defined in a sphere, being its radius less or equal to the image sphere radius. 
EquiConvs can also be seen as a general model for spherical Atrous Convolutions \cite{chen2018deeplab,chen2017rethinking} where the kernel size is what we call resolution, and the rate is the field of view of the kernel divided by the resolution. An example of the differences of \shortConvs{} by modifiying {\color{cAlphas}$\alpha$} and {\color{cRes}$r$} can be seen in Figure~\ref{fig:sphericalatrous}.

\subsection{\shortConvs{} Details}

In \cite{dai2017deformable}, they introduce deformable convolutions by learning additional offsets from the preceding feature maps. Offsets are added to the regular kernel locations in the Standard Convolution enabling free form deformation of the kernel. 

\begin{figure}
\begin{center}
\includegraphics[width=\linewidth]{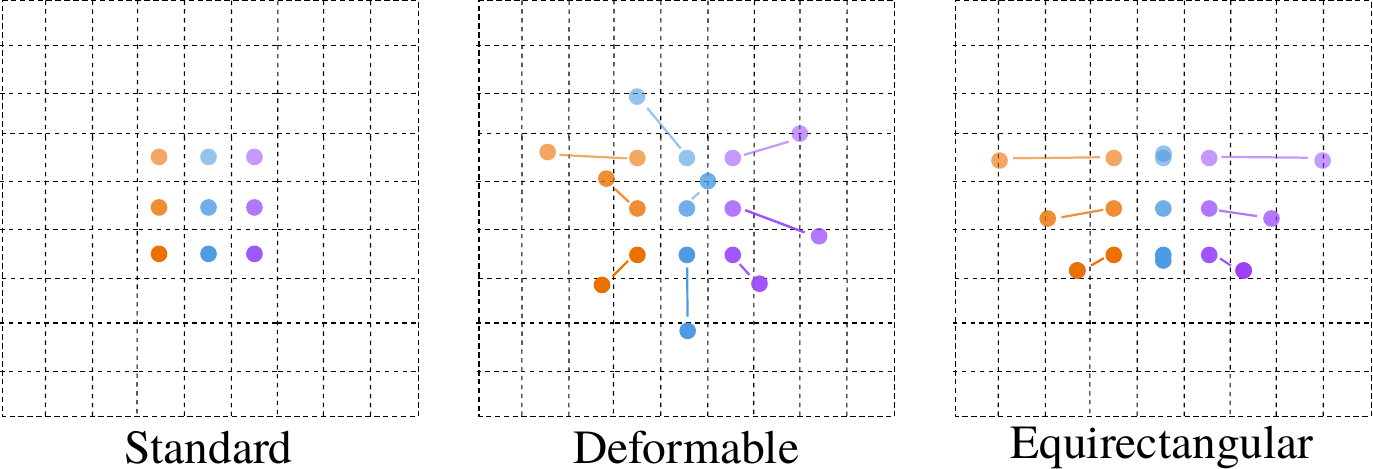}
\caption{\label{fig:kerneloffsets} \textbf{Effect of offsets on a $3\times3$ kernel}. Left: Regular kernel in Standard Convolution. Center: Deformable kernel in \cite{dai2017deformable}. Right: Spherical surface patch in \shortConvs{}.}
\end{center}
\vspace{-6mm}
\end{figure}

Inspired by this work, we deform the shape of the kernels according to the geometrical priors of the equirectangular image projection. To do that, we generate offsets that are not learned but fixed given the spherical distortion model and constant over the same horizontal locations. Here, we describe how to obtain the distorted pixel locations from the original ones.

\begin{figure}
\centering
\includegraphics[width=1\linewidth]{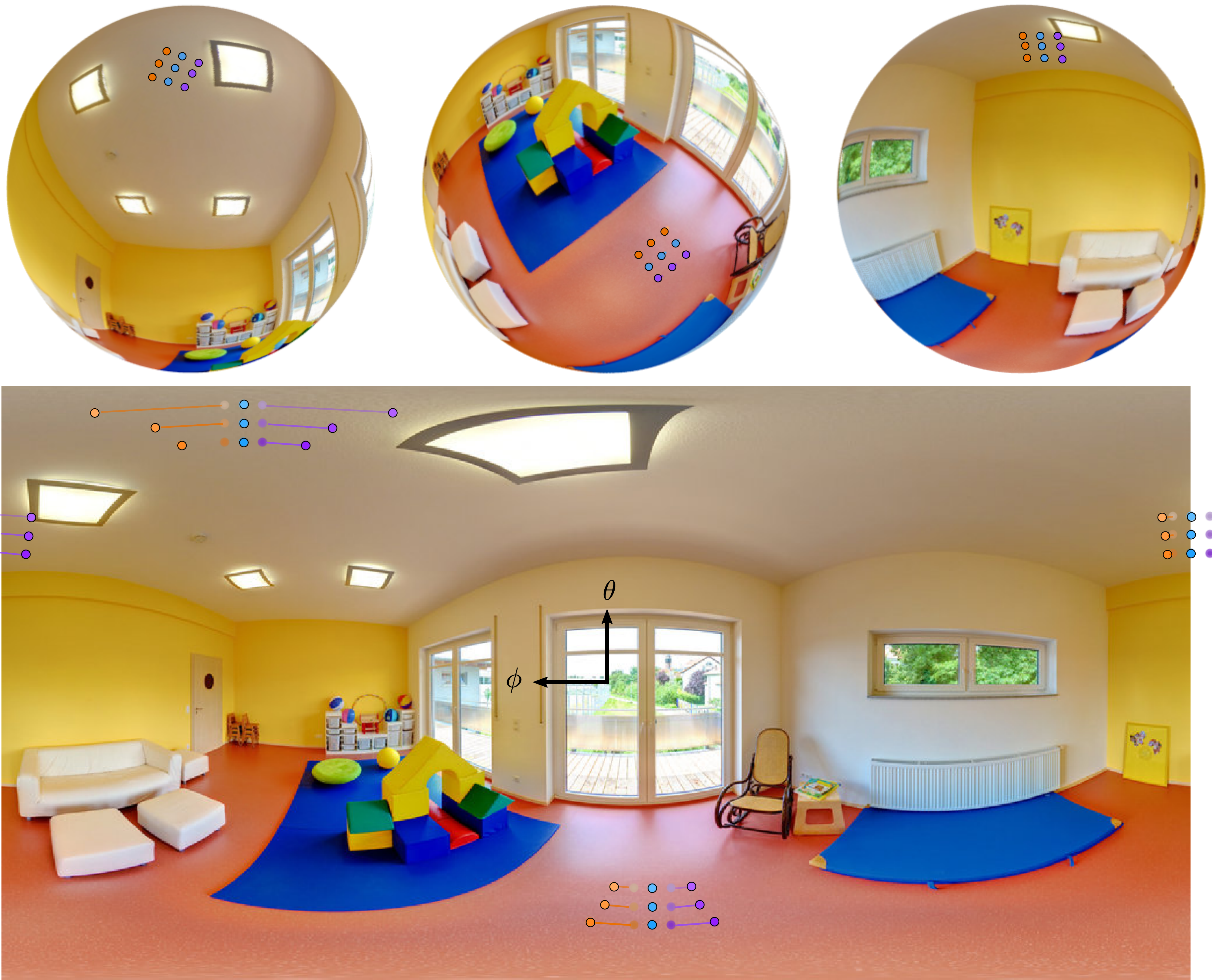}
\caption{\label{fig:convsSph} \textbf{\shortConvs{} on spherical images.} We show three kernel positions to highlight the differences between the offsets. As we approach to the poles (larger $\color{cTheta}\theta$ angles) the deformation of the kernel on the equirectangular image is bigger, in order to reproduce a regular kernel on the sphere surface. Additionally, with \shortConvs{}, we do not use padding when the kernel is on the border of the image since offsets take the points to their correct position on the other side of the $360^\circ$ image.}
\vspace{-3mm}
\end{figure}

Let us define $(u_{0,0},v_{0,0})$ as the pixel location on the equirectangular image where we apply the convolution operation (\emph{i.e.} the image coordinate where the center of the kernel is located). First, we define the coordinates for every element in the kernel and afterwards we rotate them to the point of the sphere where the kernel is being applied. We define each point of the kernel as
\begin{equation}
    \hat{p}_{ij}=  
    \begin{bmatrix}
        \hat{x}_{ij}\\
        \hat{y}_{ij}\\
        \hat{z}_{ij}
    \end{bmatrix}=\begin{bmatrix}
        i\\
        j\\
        d
    \end{bmatrix},
\end{equation}
where $i$ and $j$ are integers in the range $[-\frac{{\color{cRes}r}-1}{2},\frac{{\color{cRes}r}-1}{2}]$ and  $d$ is the distance from the center of the sphere to the kernel grid. In order to cover the field of view ${\color{cAlphas}\alpha}$,
\begin{equation}
d = \frac{{\color{cRes}r}}{2\tan(\frac{{\color{cAlphas}\alpha}}{2})}.
\end{equation}
We project each point into the sphere surface by normalizing the vectors, and rotate them to align the kernel center to the point where the kernel is applied.
\begin{equation}
    p_{ij} = \begin{bmatrix}
        {x}_{ij}\\
        {y}_{ij}\\
        {z}_{ij}
    \end{bmatrix}=R_y({\color{cPhi}\phi_{0,0}})R_x({\color{cTheta}\theta_{0,0}})\frac{\hat{p}_{ij}}{|\hat{p}_{ij}|},
\end{equation}
where $R_a(\beta)$ stands for a rotation matrix of an angle $\beta$ around the $a$ axis. ${\color{cPhi}\phi_{0,0}}$ and ${\color{cTheta}\theta_{0,0}}$ are the spherical angles of the center of the kernel --see Figure~\ref{fig:equiconv}, and are defined as 
\begin{equation}
{\color{cPhi}\phi_{0,0}} = (u_{0,0}-\dfrac{W}{2})\dfrac{2\pi}{W} \quad ;\quad  {\color{cTheta}\theta_{0,0}}  = -(v_{0,0}-\dfrac{H}{2})\dfrac{\pi}{H},
\end{equation}
where $W$ and $H$ are, respectively, the width and height of the equirectangular image in pixels. Finally, the rest of elements are back-projected to the equirectangular image domain. 
First, we convert the unit sphere coordinates to latitude and longitude angles:
\begin{equation}
\phi_{ij} =  \arctan{(\dfrac{x_{ij}}{z_{ij}})} \quad ;\quad \theta_{ij}  = \arcsin{(y_{ij})}.
\end{equation}
And then, to the original 2D equirectangular image domain:
\begin{equation}
u_{ij}  = (\dfrac{\phi_{ij} }{2\pi} + \dfrac{1}{2})W
\quad ;\quad 
v_{ij}  = (-\dfrac{\theta_{ij} }{\pi} + \dfrac{1}{2})H.
\end{equation}
In Figure~\ref{fig:kerneloffsets} we show how these offsets are applied to a regular kernel; and in Figure~\ref{fig:convsSph} three kernel samples on the spherical and on the equirectangular images.

\section{Experiments}

\begin{table*}[t!]
\scriptsize
\centering
\begin{tabular}{@{\extracolsep{4pt}}ccc ccccc c ccccc}
\multicolumn{3}{c}{}&\multicolumn{5}{c}{Edges}&&\multicolumn{5}{c}{Corners}\\
\cline{4-8} \cline{10-14}
\textbf{Conv. Type} & \textbf{IntPred}& \textbf{Edges}& $IoU$ & $Acc$ & $P$ & $R$ & $F_1$ && $IoU$ & $Acc$ & $P$ & $R$ & $F_1$\\
\hline
 & & & $:1$ & $:1$ &$:1$&$:1$&$:1$ && $:1$ & $:1$ &$:1$&$:1$&$:1$\\
StdConvs & {\color{cRes}-} & {\color{cRes}-} & - & - & - & - & - && $0.433$ & $0.971$ & $0.802$ & $0.484$ & $0.600$\\
 
StdConvs & {\color{cRes}-} & {\color{cTheta}\checkmark} &$0.564$ & $0.926$ & $0.751$ & $0.681$ & $0.713$ && $0.453$ & $0.973$ & $0.850$ & $0.493$ & $0.621$ \\

StdConvs & {\color{cTheta}\checkmark} & {\color{cTheta}\checkmark} &$\textbf{0.588}$ & $\textbf{0.933}$ & $\textbf{0.782}$ & $\textbf{0.691}$ & $\textbf{0.733}$ & & $\textbf{0.465}$ & $\textbf{0.974}$ & $\textbf{0.872}$ & $\textbf{0.498}$ & $\textbf{0.632}$

\Bstrut\\
\hline 
\shortConvs & {\color{cRes}-} & {\color{cRes}-} & - & - & - & - & - && $0.437$ & $0.970$ & $0.784$ & $0.496$ & $0.604$ \Tstrut\\

\shortConvs & {\color{cRes}-} & {\color{cTheta}\checkmark} & $0.548$ & $0.920$ & $0.718$ & $\textbf{0.686}$ & $0.700$ && $0.444$ & $0.972$ & $0.822$ & $\textbf{0.491}$ & $0.611$\\

\shortConvs & {\color{cTheta}\checkmark} & {\color{cTheta}\checkmark} & $\textbf{0.575}$ & $\textbf{0.931}$ & $\textbf{0.789}$ & $0.667$ & $\textbf{0.722}$ && $\textbf{0.460}$ & $\textbf{0.974}$ & $\textbf{0.887}$ & $0.488$ & $\textbf{0.627}$\\

\cline{4-8} \cline{10-14}
\multicolumn{3}{c}{}&\multicolumn{5}{c}{\btb{}}&&\multicolumn{5}{c}{\btb{}}\\
\end{tabular}
\caption{\label{tab:ablationStd}\textbf{Ablation study on SUN360 dataset.} We show results for both Standard Convolutions (StdConvs) and our proposed \longConvs{} (\shortConvs{}) with some modifications: Using or not intermediate predictions (IntPred) in the decoder and edge map predictions (Edges).}
\vspace{-5mm}
\end{table*}

We present a set of experiments to evaluate CFL using both Standard Convolutions (StdConvs) and the proposed \longConvs{} (\shortConvs{}). We do not only analyze how well it predicts edge and corner maps, but also the impact of each algorithmic component through ablation studies. We report the performance of our proposal in two different datasets, and show qualitative 2D and 3D models of different indoor scenes.

\subsection{Datasets}
We use two public datasets that comprise several indoor scenes, SUN360 \cite{Xiao2012} and Stanford (2D-3D-S) \cite{2017arXiv170201105A} in equirectangular projection (360$^\circ$). The former is used for ablation studies, and both are used for comparison against several state-of-the-art baselines. 

\noindent\textbf{SUN360 \cite{Xiao2012}}: We use $\thicksim$500 bedroom and livingroom panoramas from this dataset labeled by Zhang \textit{et al.} \cite{PanoContext}. %
We use these labels but, since all panoramas were labeled as box-type rooms, we hand-label and substitute 35 panoramas representing more faithfully the actual shapes of the rooms. We split the raw dataset in 85$\%$ training scenes and 15$\%$ test scenes randomly by making sure that there were rooms of more than 4 walls in both partitions.

\noindent\textbf{Stanford 2D-3D-S \cite{2017arXiv170201105A}}: This dataset contains more challenging scenarios like cluttered laboratories or corridors. In \cite{zou2018layoutnet}, they use areas 1, 2, 4, 6 for training, and area 5 for testing. For our experiments we use same partitions and the ground truth provided by them.

\vspace{-1mm}
\subsection{Implementation details}
\vspace{-1mm}
The input to the network is a single panoramic RGB image of resolution $256\times128$. 
The outputs are, on the one hand, the room layout edge map and on the other hand, the corner map, both of them at resolution $128\times64$. 
A widely used strategy to improve generalization of neural networks is data augmentation. We apply random erasing, horizontal mirroring as well as horizontal rotation from $0^\circ$ to $360^\circ$ of input images during training. The weights are all initialized using ResNet-50 \cite{he2016deep} trained on ImageNet \cite{russakovsky2015imagenet}. For \textit{CFL EquiConvs} we use the same kernel resolutions and field of views as in ResNet-50. This means that for a standard 3$\times$3 kernel applied to a W$\times$H feature map, {\color{cRes}$r$}$=3$ and {\color{cAlphas}$\alpha$}$= ${\color{cRes}$r$}$ \frac{fov}{W}$, where $fov=360^\circ$ for panoramas.
We minimize the cross-entropy loss using Adam \cite{kingma2014adam}, regularized by penalizing the loss with the sum of the L2 of all weights. The initial learning rate is $2.5e^{-4}$ and is exponentially decayed by a rate of $0.995$ every epoch. We apply a dropout rate of $0.3$.%

The network is implemented using TensorFlow \cite{abadi2016tensorflow} and trained and tested in a NVIDIA Titan X. The training time for StdConvs is around $1$ hour and the test time is $0.31$ seconds per image. For \shortConvs{}, training takes $3$ hours and test around $3.32$ seconds per image. %

\subsection{FCN evaluation}

We measure the quality of our predicted probability maps using five standard metrics: intersection over union of predicted corner/edge pixels \textit{IoU}, precision \textit{P}, recall \textit{R}, F1 Score \textit{$F_{1}$} and accuracy \textit{Acc}.
Table~\ref{tab:ablationStd} summarizes our results and allows us to answer the following questions: 

\noindent\textbf{What are the effects of different convolutions?} 
As one would expect, \shortConvs{}, aware of the distortion model,  learn in a non-distorted generic feature space achieving accurate predictions, like StdConvs on conventional images \cite{RoomNet}. However and counterintuitively, StdConvs, ignoring the distortion model, rely on image patterns that this generates obtaining similar performance -- see Table~\ref{tab:ablationStd}. Distortion understanding, nonetheless, gives the network other advantages. While StdConvs learn strong bias correlation between features and distortion patterns (\textit{e.g.} ceiling line on the top of the image or clutter in the mid-bottom), EquiConvs are invariant to that. For this reason, the performance of \shortConvs{} does not degrade when varying the camera 6DOF pose -- see Section~\ref{RobustSect}. Additionally, EquiConvs allow a more direct use of networks pre-trained on conventional images. Specifically, this translates into a faster convergence, which is desirable as, to date, $360^\circ$ datasets contain far less images than datasets with conventional images. Moreover Tateno \textit{et al.} demonstrate in their recent work \cite{tateno2018distortion} that other tasks like depth prediction, panoramic monocular SLAM, panoramic semantic segmentation and panoramic style transfer can also benefit from this type of convolutions.

\noindent\textbf{How can we refine predictions?} There are some techniques that we can use in order to obtain more accurate and refined predictions. Here, we make pyramid preliminary predictions in the decoder and iteratively refine them, by feeding them back to the network, until the final prediction. Also, although we only use the corner map to recover the layout of the room, we train the network to additionally predict edge maps as an auxiliary task. This is another representation of the same task that ensures that the network learns to exploit the relationship between both outputs, \textit{i.e.}, the network learns how edges intersect between them generating the corners. The improvement is shown in the Table~\ref{tab:ablationStd}.

\begin{table*}[ht!]
\scriptsize
\centering
\begin{tabular}{@{\extracolsep{3pt}}c c ccc c ccc}
\multicolumn{2}{c}{}&\multicolumn{3}{c}{Edges}&&\multicolumn{3}{c}{Corners}\\
\cline{3-6} \cline{7-9}
 &  &  $F_{1}$ & $Acc$ & $IoU$ && $F_{1}$ & $Acc$ & $IoU$ \\
\hline
 & & $\%$ & $\%$ & $\%$ && $\%$ & $\%$ & $\%$ \\
\multirow{2}{*}{\color{trasla}\textbf{\small Translation} (-0.3h:+0.3h)} & StdConvs & $63.00 \pm 5.85$ & $89.70 \pm 1.89$ & $46.25 \pm 6.20$ && $43.97 \pm 5.70$ & $97.79 \pm 0.25$ & $28.35 \pm 4.71$\\
 & \shortConvs & $\textbf{64.25} \pm \textbf{2.36}$ & $\textbf{90.16} \pm \textbf{0.8}$ & $\textbf{47.37} \pm \textbf{2.57}$ && $\textbf{44.75} \pm \textbf{5.34}$ & $\textbf{97.88} \pm \textbf{0.20}$ & $\textbf{28.97} \pm \textbf{4.34}$\\ 
\noalign{\smallskip}
\hline
\noalign{\smallskip}
\multirow{2}{*}{\color{rota}\textbf{\small Rotation} ($-30^\circ$:$+30^\circ$)} & StdConvs & $54.99 \pm 11.8$ & $86.83 \pm 4.7$  & $38.88 \pm 11.7$ && $33.47 \pm 12.9$ & $97.38 \pm 0.6$  & $20.84 \pm 9.7$\\
 & \shortConvs & $\textbf{59.51} \pm \textbf{9.2}$ & $\textbf{88.64} \pm \textbf{3.5}$ & $\textbf{42.97} \pm \textbf{9.4}$ && $\textbf{35.82} \pm \textbf{12.4}$ & $\textbf{97.66} \pm \textbf{0.4}$ & $\textbf{22.53} \pm \textbf{9.5}$\\
\end{tabular}
\caption{\label{tab:Robustness}\textbf{Robustness analysis}. Values represent the mean value (\btb) $\pm$ standard deviation (\stb). We apply two types of transformations to the panoramas: {\color{trasla}\textbf{translations}} in $y$ dependant on the room height, $h$, and {\color{rota}\textbf{rotations}} in $x$. We do not use these images for training but just for testing in order to show the generalization capabilities of \shortConvs.}
\vspace{-5mm}
\end{table*}

\begin{figure}
\begin{center}
     \includegraphics[width=1\linewidth]{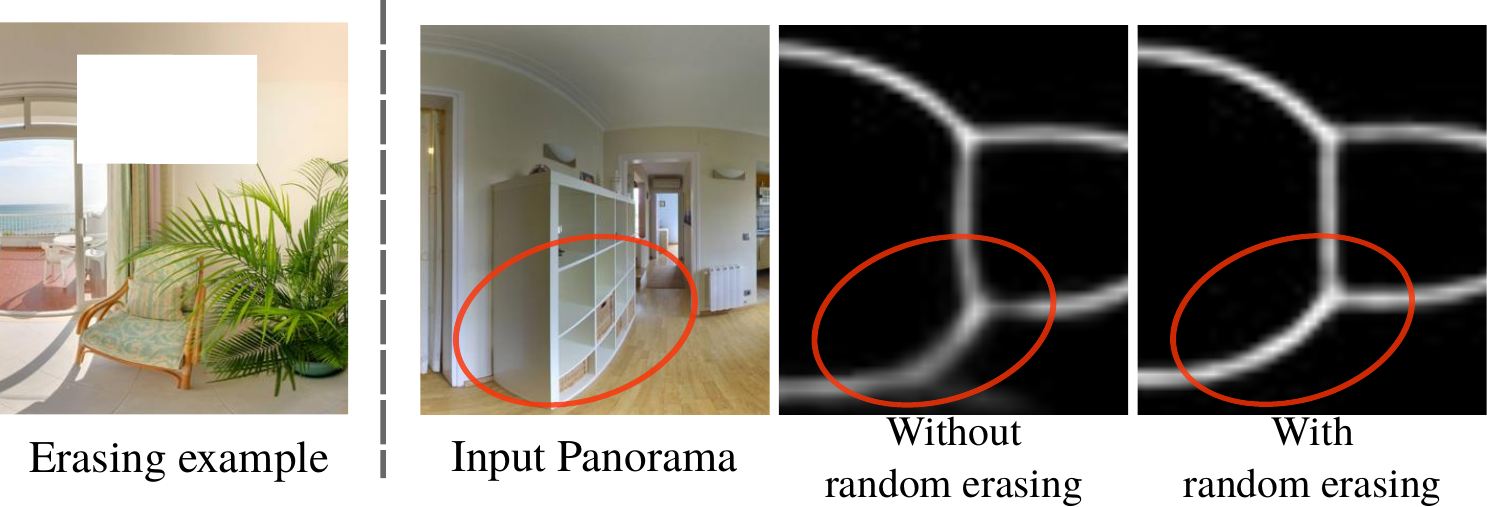}
\end{center}
\vspace{-2mm}
\caption{\textbf{Augmenting the data with virtual occlusions.} Left: Image with erased pixels. Right: Input panorama and predictions without and with pixel erasing. Notice the improvement by random erasing.}
\label{fig:occlusions}
\vspace{-3mm}
\end{figure}

\noindent\textbf{How can we deal with occlusions?} 
We do Random Erasing Data Augmentation. This operation randomly selects rectangles in the training images and removes its content, generating various levels of virtual occlusion. In this manner we simulate real situations where objects in the scene occlude the corners of the room layout, and force the network to learn context-aware features to overcome this challenging situation. Figure~\ref{fig:occlusions} illustrates this strategy with an example.

\begin{figure}
\centering
\includegraphics[width=1\linewidth]{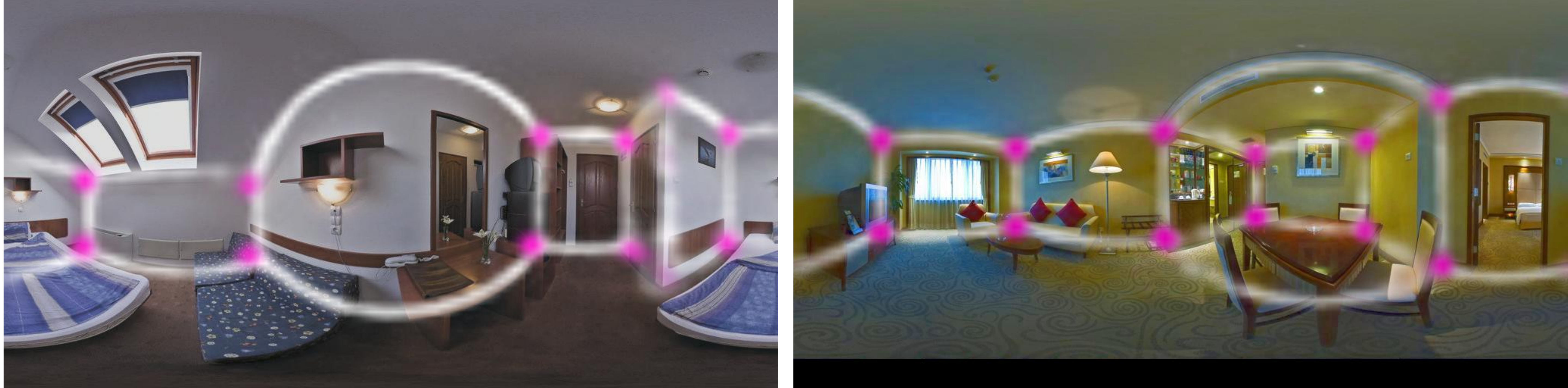}
\caption{\label{fig:nonMan} \textbf{Relaxation of assumptions.} The figure shows two CFL predictions of non-Manhattan/not box-like rooms.} 
\vspace{-5mm}
\end{figure}

\noindent\textbf{Is it possible to relax the scene assumptions while keeping a good performance?} Our end-to-end approach overcomes the Manhattan assumption as well as the box-type simplification (four-walls rooms).
On the one hand, although we label some panoramas more accurately to their actual shape, we still have a largely unbalanced dataset. We address this problem by choosing a batch size of $16$ and forcing it to always include one non-box sample. This favors the learning of more complex rooms despite having few examples.
On the other hand, while recent works \cite{zou2018layoutnet,fernandez2018layouts,PanoContext} use pre-computed vanishing points and posterior optimizations, here we directly obtain the corner coordinates from the FCN output without applying geometric constraints. In Figure~\ref{fig:nonMan} we show two examples where CFL predicts more than $4$ walls. Notice also the non-Manhattan ceiling in the left image.

\subsection{Robustness analysis}
\label{RobustSect}
\begin{figure}[t!]
\begin{center}
{\includegraphics[width=0.46\linewidth,cfbox=trasla 2pt 0pt 0pt]{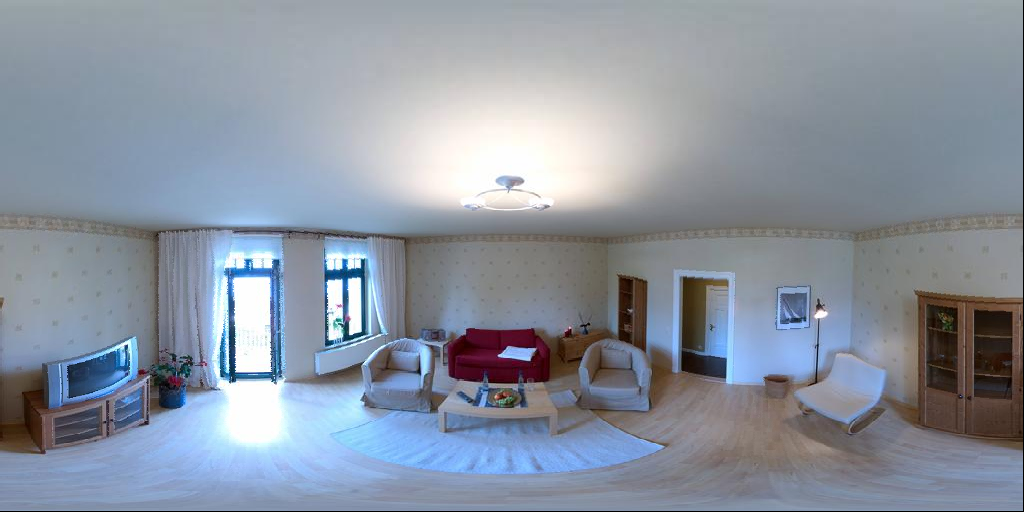}}\hspace*{0.005\linewidth}
{\includegraphics[width=0.46\linewidth,cfbox=rota 2pt 0pt 0pt]{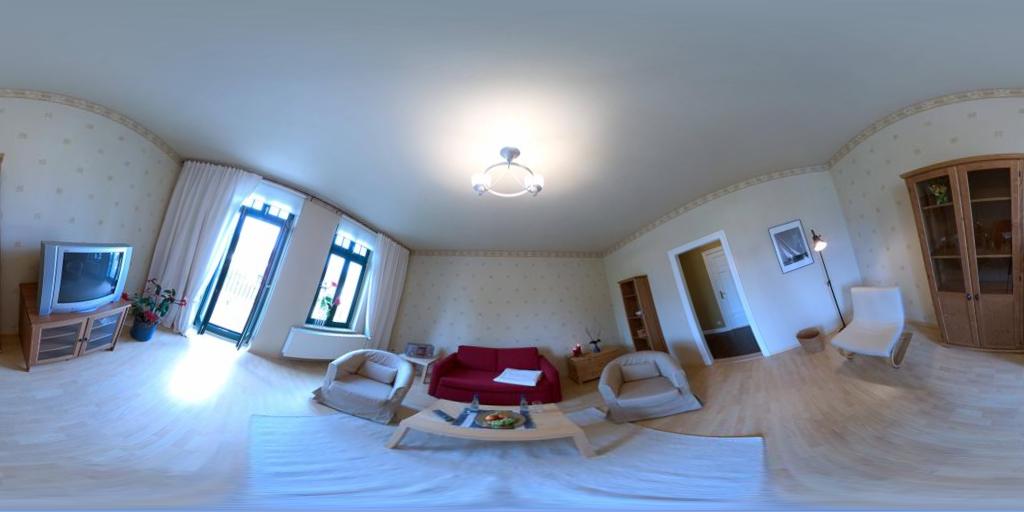}}
\caption{\label{fig:rot&trans} \textbf{Synthetic images for robustness analysis.} Here we show two examples of panoramas generated with upward {\color{trasla}\textbf{translation}} in $y$ and {\color{rota}\textbf{rotation}} in $x$ respectively.}
\end{center}
\vspace{-9mm}
\end{figure}

With the motivation of exploiting the potential of EquiConvs, we test our model with previously unseen images where the camera viewpoint is different from that in the training set. The distortion in equirectangular projection is location dependent, specifically, it depends on the polar angle {\color{cTheta}$\theta$}. Since EquiConvs are invariant to this distortion, it is interesting to see how modifications in the camera extrinsic parameters ({\color{trasla}\textbf{translation}} and {\color{rota}\textbf{rotation}}) affect the model performance using EquiConvs against StdConvs.  When we generate translations over the vertical axis and rotations, the shape of the layout is modified by the distortion, losing its characteristic pattern (which StdConvs use in its favor).%
Since standard datasets have a strong bias when referring to camera pose and rotation, we synthetically render these transformations along our test set. The {\color{rota}\textbf{rotation}} is trivial as we work on the spherical domain. As the complete 3D dense model of the rooms is not available, the {\color{trasla}\textbf{translation}} simulation is performed by using the existing information, ignoring occlusions produced by viewpoint changes. Nevertheless, as we do not work with wide translations the effect is minimal and images are realistic enough to prove the point we want to highlight (see Figure \ref{fig:rot&trans}). %

For both experiments, we uniformly sample from a minimum to a maximum transformation and calculate the mean and standard deviation for all the metrics. What we see in Table~\ref{tab:Robustness} is that we obtain higher mean values while smaller standard deviation by using EquiConvs. This means that this EquiConvs make the model more robust and generalizable to real life situations, not covered in the datasets, \eg panoramas taken by hand, drones or small robots. This effect is highlighted especially in the evaluation of the edges since it is their appearance that is highly modified by these changes of the camera.

\subsection{3D Layout comparison}%

We evaluate our layout predictions using three standard metrics, 3D intersection over union $3DIoU$, corner error $CE$ and pixel error $PE$, and compare ourselves against four approaches from the state of the art \cite{PanoContext,zou2018layoutnet,fernandez2018layouts, yang2018dula}. Pano2CAD \cite{Pano2cad} has no source code available nor evaluation of layouts, making direct comparison difficult. 
The pixel error metric given by \cite{zou2018layoutnet} only distinguishes between ceiling, floor and walls, $PE^{SS}$. Instead our proposed segmented mask distinguish between ceiling, floor and each wall separately, $PE^{CS}$, which is more informative since it also has into account errors in wall-wall boundaries.
For all experiments, only SUN360 dataset is used for training.
Table~\ref{tab:Layoutresults} shows the performance of our proposal testing on both datasets, SUN360 and Stanford 2D-3D. Results are averaged across all images. It can be seen that our approach outperforms the state of the art clearly, in all the metrics.

It is worth mentioning that our approach, not only obtains better accuracy but also it recovers shapes more faithful to the real ones, since it can handle non box-type room designs with few training examples.
In Table~\ref{table:CTime} we show that, apart from achieving better localization of layout boundaries and corners, our end-to-end approach is much faster. Our full method with \shortConvs{} takes $3.47$ seconds to process one room and with StdConvs just $0.46$ seconds, which is a major advantage considering the aforementioned applications of layout recovery need to be real-time (robot navigation, AR/VR).%

\begin{table}
\footnotesize{}%
\centering
\begin{tabular}{@{\extracolsep{0pt}}ccc ccc}
\multicolumn{6}{c}{}\\ 
\textbf{Test} & \textbf{Method}& $3DIoU$ & $CE$ & $PE^{SS}$ & $PE^{CS}$\\
\hline
 & & $\%$ & $\%$ & $\%$ & $\%$\\
\multirow{5}{*}{SUN360} & PanoContext \cite{PanoContext}&  $67.22$ & $1.60$  & $4.55$ & $10.34$ \\
&Fernandez \cite{fernandez2018layouts} & - & - & - & $7.26$\\
&LayoutNet \cite{zou2018layoutnet} & $74.48$ & $1.06$ & $3.34$ & -\\
&DuLa-Net \cite{yang2018dula} & $77.42$ & - & - & -\\
& CFL StdConvs  &  $\textbf{78.79}$ & $0.79$ & $\textbf{2.49}$ & $\textbf{3.33}$   \\ 
& CFL \shortConvs   &  $77.63$ & $\textbf{0.78}$ & $2.64$ & $3.35$   \\ 
\noalign{\smallskip}
\hline
\noalign{\smallskip}
\multirow{3}{*}{Std.2D3D} & Fernandez \cite{fernandez2018layouts} & - & - & - & $12.1$\\
& CFL StdConvs  &  $65.13$ & $\textbf{1.44}$ & $\textbf{4.75}$ & $\textbf{6.05}$ \\ 
& CFL \shortConvs   &  $\textbf{65.23}$ & $1.64$ & $5.52$ & $7.11$\\
\cline{4-6}
\multicolumn{2}{c}{}&\multicolumn{1}{c}{}&\multicolumn{3}{c}{\stb{}}\\
\end{tabular}
\caption{\label{tab:Layoutresults} \textbf{Layout results} on both datasets, training on SUN360 data. \textit{SS}: Simple Segmentation (3 categories): ceiling, floor and walls \cite{zou2018layoutnet}. \textit{CS}: Complete Segmentation: ceiling, floor, wall$_{1}$,..., wall$_{n}$ \cite{fernandez2018layouts}. Observe how our method outperforms all the baselines in all the metrics.}
\vspace{-2mm}
\end{table}

\begin{table}
\small
\begin{center}
\begin{tabular}{cc}
\hline\noalign{\smallskip}
\textbf{Method} &\textbf{Computation Time (s)} \\
\noalign{\smallskip}
\hline
\noalign{\smallskip}
PanoContext \cite{PanoContext} & $>300$\\
LayoutNet \cite{zou2018layoutnet} &  $44.73$ \\
DuLa-Net \cite{yang2018dula} &  $13.43$ \\
CFL \shortConvs  &  $3.47$\\ 
CFL StdConvs  &  $\textbf{0.46}$\\ 
\hline
\end{tabular}

\caption{\textbf{Average computing time per image.} Every approach is evaluated using NVIDIA Titan X and Intel Xeon 3.5 GHz (6 cores) except DuLa-Net, evaluated using NVIDIA 1080ti GPU. Our end-to-end method is more than 100 times faster than other methods.}
\label{table:CTime}
\end{center}
\vspace{-7mm}
\end{table}

\begin{figure}
\begin{center}
     \includegraphics[width=1\linewidth]{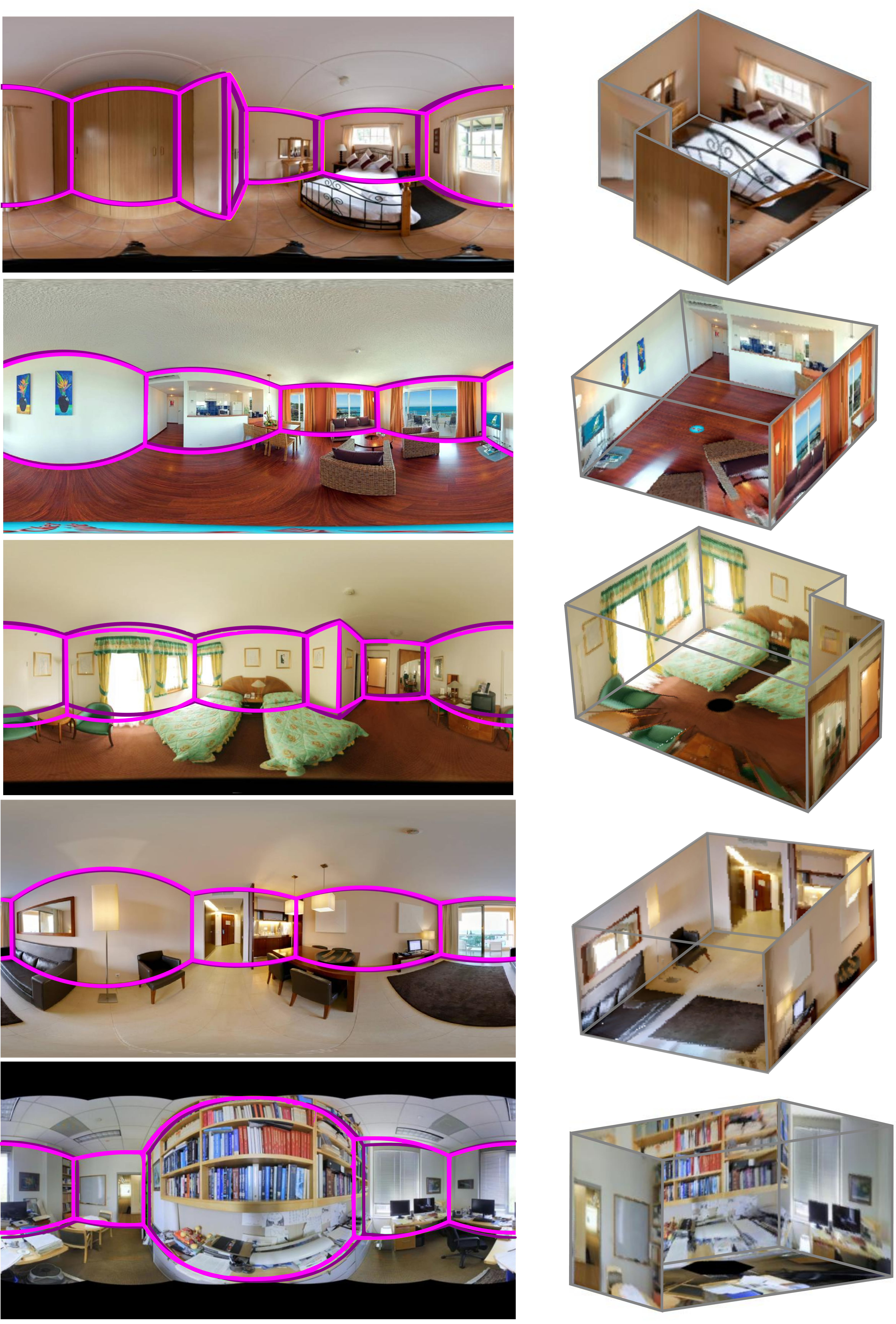}
\end{center}
\vspace{-2mm}
\caption{Layout predictions ({\color{magenta}light magenta}) and ground truth ({\color{dmagenta}dark magenta}) for \textbf{complex room geometries}.}
\label{fig:experi2}
\vspace{-5mm}
\end{figure}

\section{Conclusions}

In this work we present \shortMethod{}, the first end-to-end algorithm for layout recovery in $360^\circ$ images. %
Our experimental results demonstrate that our predicted layouts are clearly more accurate than the state of the art. Additionally, the removal of extra pre- and post-processing stages makes our method much faster than other works. Finally, being entirely data-driven removes the geometric assumptions that are commonly used in the state of the art and limits their usability in complex geometries. We present two different variants of \shortMethod{}. The first one, implemented using Standard Convolutions, reduces the computation in 100 times and it is very suitable for images taken with a tripod. The second one uses our proposed implementation of \longConvs{} that adapt their shape to the equirectangular projection of the spherical image. This proves to be more robust to translations and rotations of the camera making it ideal for panoramas taken by a hand-held camera.

{
\small
\vspace{1ex}
\noindent\textbf{Acknowledgement: }This project was in part funded by the Spanish government (DPI2015-65962-R, DPI2015-67275), the Regional Council of Bourgogne-Franche-Comt\'e (2017-9201AAO048S01342) and the Aragon government (DGA-T45\_17R/FSE). We also thank Nvidia for their Titan X and Xp donation. Also, we would like to acknowledge Jesus Bermudez-Cameo for his valuable discussions and alpha testing.
}

{\small
\bibliographystyle{ieee}
\bibliography{cflbib}

\begin{thebibliography}{10}\itemsep=-1pt

\bibitem{abadi2016tensorflow}
M.~Abadi, P.~Barham, J.~Chen, Z.~Chen, A.~Davis, J.~Dean, M.~Devin,
  S.~Ghemawat, G.~Irving, M.~Isard, et~al.
\newblock Tensorflow: A system for large-scale machine learning.
\newblock In {\em OSDI}, volume~16, pages 265--283, 2016.

\bibitem{2017arXiv170201105A}
I.~{Armeni}, A.~{Sax}, A.~R. {Zamir}, and S.~{Savarese}.
\newblock {Joint 2D-3D-Semantic Data for Indoor Scene Understanding}.
\newblock {\em ArXiv}, Feb. 2017.

\bibitem{bao2011toward}
S.~Y. Bao, M.~Sun, and S.~Savarese.
\newblock Toward coherent object detection and scene layout understanding.
\newblock {\em Image and Vision Computing}, 29(9):569--579, 2011.

\bibitem{chen2018deeplab}
L.-C. Chen, G.~Papandreou, I.~Kokkinos, K.~Murphy, and A.~L. Yuille.
\newblock Deeplab: Semantic image segmentation with deep convolutional nets,
  atrous convolution, and fully connected crfs.
\newblock {\em IEEE transactions on pattern analysis and machine intelligence},
  40(4):834--848, 2018.

\bibitem{chen2017rethinking}
L.-C. Chen, G.~Papandreou, F.~Schroff, and H.~Adam.
\newblock Rethinking atrous convolution for semantic image segmentation.
\newblock {\em arXiv preprint arXiv:1706.05587}, 2017.

\bibitem{cohen2018spherical}
T.~S. Cohen, M.~Geiger, J.~K{\"o}hler, and M.~Welling.
\newblock Spherical cnns.
\newblock {\em arXiv preprint arXiv:1801.10130}, 2018.

\bibitem{dai2017deformable}
J.~Dai, H.~Qi, Y.~Xiong, Y.~Li, G.~Zhang, H.~Hu, and Y.~Wei.
\newblock Deformable convolutional networks.
\newblock {\em CoRR, abs/1703.06211}, 1(2):3, 2017.

\bibitem{dasgupta2016delay}
S.~Dasgupta, K.~Fang, K.~Chen, and S.~Savarese.
\newblock Delay: Robust spatial layout estimation for cluttered indoor scenes.
\newblock In {\em IEEE Conference on Computer Vision and Pattern Recognition},
  pages 616--624, 2016.

\bibitem{Delage2006}
E.~Delage, H.~Lee, and A.~Y. Ng.
\newblock A dynamic bayesian network model for autonomous 3{D} reconstruction
  from a single indoor image.
\newblock In {\em IEEE Computer Society Conference on Computer Vision and
  Pattern Recognition}, volume~2, pages 2418--2428, 2006.

\bibitem{dosovitskiy2015flownet}
A.~Dosovitskiy, P.~Fischer, E.~Ilg, P.~Hausser, C.~Hazirbas, V.~Golkov,
  P.~van~der Smagt, D.~Cremers, and T.~Brox.
\newblock Flownet: Learning optical flow with convolutional networks.
\newblock In {\em Proceedings of the IEEE International Conference on Computer
  Vision}, pages 2758--2766, 2015.

\bibitem{Eigen2015}
D.~Eigen and R.~Fergus.
\newblock Predicting depth, surface normals and semantic labels with a common
  multi-scale convolutional architecture.
\newblock In {\em IEEE International Conference on Computer Vision}, pages
  2650--2658, 2015.

\bibitem{fernandez2018panoroom}
C.~Fernandez-Labrador, J.~M. Facil, A.~Perez-Yus, C.~Demonceaux, and J.~J.
  Guerrero.
\newblock Panoroom: From the sphere to the 3d layout.
\newblock {\em arXiv preprint arXiv:1808.09879}, 2018.

\bibitem{fernandez2018layouts}
C.~Fernandez-Labrador, A.~Perez-Yus, G.~Lopez-Nicolas, and J.~J. Guerrero.
\newblock Layouts from panoramic images with geometry and deep learning.
\newblock {\em IEEE Robotics and Automation Letters}, 3(4):3153--3160, 2018.

\bibitem{flint2011manhattan}
A.~Flint, D.~Murray, and I.~Reid.
\newblock Manhattan scene understanding using monocular, stereo, and 3d
  features.
\newblock In {\em Computer Vision (ICCV), 2011 IEEE International Conference
  on}, pages 2228--2235. IEEE, 2011.

\bibitem{fouhey2014people}
D.~F. Fouhey, V.~Delaitre, A.~Gupta, A.~A. Efros, I.~Laptev, and J.~Sivic.
\newblock People watching: Human actions as a cue for single view geometry.
\newblock {\em International journal of computer vision}, 110(3):259--274,
  2014.

\bibitem{he2016deep}
K.~He, X.~Zhang, S.~Ren, and J.~Sun.
\newblock Deep residual learning for image recognition.
\newblock In {\em Proceedings IEEE Conference on Computer Vision and Pattern
  Recognition}, pages 770--778, 2016.

\bibitem{Hedau2009}
V.~Hedau, D.~Hoiem, and D.~Forsyth.
\newblock Recovering the spatial layout of cluttered rooms.
\newblock In {\em IEEE International Conference on Computer Vision}, pages
  1849--1856, 2009.

\bibitem{karsch2011rendering}
K.~Karsch, V.~Hedau, D.~Forsyth, and D.~Hoiem.
\newblock Rendering synthetic objects into legacy photographs.
\newblock {\em ACM Transactions on Graphics (TOG)}, 30(6):157, 2011.

\bibitem{kingma2014adam}
D.~P. Kingma and J.~Ba.
\newblock Adam: A method for stochastic optimization.
\newblock {\em arXiv preprint arXiv:1412.6980}, 2014.

\bibitem{RoomNet}
C.~Lee, V.~Badrinarayanan, T.~Malisiewicz, and A.~Rabinovich.
\newblock {RoomNet}: End-to-end room layout estimation.
\newblock In {\em IEEE International Conference on Computer Vision}, 2017.

\bibitem{Lee2009}
D.~C. Lee, M.~Hebert, and T.~Kanade.
\newblock Geometric reasoning for single image structure recovery.
\newblock In {\em IEEE Conference on Computer Vision and Pattern Recognition
  (CVPR)}, pages 2136--2143, 2009.

\bibitem{Liu_2015_CVPR}
C.~Liu, A.~G. Schwing, K.~Kundu, R.~Urtasun, and S.~Fidler.
\newblock Rent3d: Floor-plan priors for monocular layout estimation.
\newblock In {\em The IEEE Conference on Computer Vision and Pattern
  Recognition (CVPR)}, June 2015.

\bibitem{Mallya:2015}
A.~Mallya and S.~Lazebnik.
\newblock Learning informative edge maps for indoor scene layout prediction.
\newblock In {\em IEEE International Conference on Computer Vision}, pages
  936--944, 2015.

\bibitem{ren2016coarse}
Y.~Ren, S.~Li, C.~Chen, and C.-C.~J. Kuo.
\newblock A coarse-to-fine indoor layout estimation (cfile) method.
\newblock In {\em Asian Conference on Computer Vision}, pages 36--51, 2016.

\bibitem{ronneberger2015u}
O.~Ronneberger, P.~Fischer, and T.~Brox.
\newblock U-net: Convolutional networks for biomedical image segmentation.
\newblock In {\em MICCAI}, pages 234--241, 2015.

\bibitem{russakovsky2015imagenet}
O.~Russakovsky, J.~Deng, H.~Su, J.~Krause, S.~Satheesh, S.~Ma, Z.~Huang,
  A.~Karpathy, A.~Khosla, M.~Bernstein, et~al.
\newblock Imagenet large scale visual recognition challenge.
\newblock {\em International Journal of Computer Vision}, 115(3):211--252,
  2015.

\bibitem{schwing2013box}
A.~G. Schwing, S.~Fidler, M.~Pollefeys, and R.~Urtasun.
\newblock Box in the box: Joint 3{D} layout and object reasoning from single
  images.
\newblock In {\em IEEE International Conference on Computer Vision}, pages
  353--360, 2013.

\bibitem{simonyan2014very}
K.~Simonyan and A.~Zisserman.
\newblock Very deep convolutional networks for large-scale image recognition.
\newblock {\em arXiv preprint arXiv:1409.1556}, 2014.

\bibitem{song2016deep}
S.~Song and J.~Xiao.
\newblock Deep sliding shapes for amodal 3d object detection in rgb-d images.
\newblock In {\em Proceedings of the IEEE Conference on Computer Vision and
  Pattern Recognition}, pages 808--816, 2016.

\bibitem{tateno2018distortion}
K.~Tateno, N.~Navab, and F.~Tombari.
\newblock Distortion-aware convolutional filters for dense prediction in
  panoramic images.
\newblock In {\em Proceedings of the European Conference on Computer Vision
  (ECCV)}, pages 707--722, 2018.

\bibitem{tsai2011real}
G.~Tsai, C.~Xu, J.~Liu, and B.~Kuipers.
\newblock Real-time indoor scene understanding using bayesian filtering with
  motion cues.
\newblock In {\em Computer Vision (ICCV), 2011 IEEE International Conference
  on}, pages 121--128. IEEE, 2011.

\bibitem{Xiao2012}
J.~Xiao, K.~Ehinger, A.~Oliva, and A.~Torralba.
\newblock Recognizing scene viewpoint using panoramic place representation.
\newblock In {\em IEEE Conference on Computer Vision and Pattern Recognition},
  pages 2695--2702, 2012.

\bibitem{Pano2cad}
J.~Xu, B.~Stenger, T.~Kerola, and T.~Tung.
\newblock {Pano2CAD}: Room layout from a single panorama image.
\newblock In {\em IEEE Winter Conference on Applications of Computer Vision},
  pages 354--362, 2017.

\bibitem{yang2018dula}
S.-T. Yang, F.-E. Wang, C.-H. Peng, P.~Wonka, M.~Sun, and H.-K. Chu.
\newblock Dula-net: A dual-projection network for estimating room layouts from
  a single rgb panorama.
\newblock {\em arXiv preprint arXiv:1811.11977}, 2018.

\bibitem{zhang2013estimating}
J.~Zhang, C.~Kan, A.~G. Schwing, and R.~Urtasun.
\newblock Estimating the 3d layout of indoor scenes and its clutter from depth
  sensors.
\newblock In {\em 2013 IEEE International Conference on Computer Vision}, pages
  1273--1280. IEEE, 2013.

\bibitem{Mallya2}
W.~Zhang, W.~Zhang, K.~Liu, and J.~Gu.
\newblock Learning to predict high-quality edge maps for room layout
  estimation.
\newblock {\em Transactions on Multimedia}, 19(5):935--943, 2017.

\bibitem{PanoContext}
Y.~Zhang, S.~Song, P.~Tan, and J.~Xiao.
\newblock Pano{C}ontext: A whole-room 3{D} context model for panoramic scene
  understanding.
\newblock In {\em European Conference on Computer Vision}, pages 668--686.
  Springer, 2014.

\bibitem{zhao2017physics}
H.~Zhao, M.~Lu, A.~Yao, Y.~Guo, Y.~Chen, and L.~Zhang.
\newblock Physics inspired optimization on semantic transfer features: An
  alternative method for room layout estimation.
\newblock {\em arXiv preprint arXiv:1707.00383}, 2017.

\bibitem{zou2018layoutnet}
C.~Zou, A.~Colburn, Q.~Shan, and D.~Hoiem.
\newblock Layoutnet: Reconstructing the 3d room layout from a single rgb image.
\newblock In {\em Proceedings IEEE Conference on Computer Vision and Pattern
  Recognition}, pages 2051--2059, 2018.

\end{thebibliography}
}

\clearpage

\end{document}